\documentclass[review]{elsarticle}
\graphicspath{ {./Figures/} }
\usepackage{hyperref}
\usepackage{float}
\usepackage{verbatim} 
\usepackage{apalike}
\restylefloat{figure}
\restylefloat{table}
\usepackage{booktabs}
\usepackage{multicol}  
\usepackage{multirow}  
\usepackage{tabu}  
\usepackage{color}
\usepackage{amsfonts}
{}
\usepackage{amsmath,graphicx}

\usepackage[figuresright]{rotating} 
\usepackage{pdflscape}

\biboptions{authoryear}

\begin{document}
\begin{frontmatter}

\title{Rethinking the Detection Head Configuration for Traffic Object Detection}

%

\author[label1]{Yi Shi}
\ead{yishi701@gmail.com}

\author[label1]{Jiang Wu}
\ead{jiangjiangbabay@gmail.com}

\author[label1]{Shixuan Zhao}
\ead{zhaosx@std.uestc.edu.cn}


\author[label1]{Gangyao Gao}
\ead{gangyaogao@gmail.com}

\author[label2]{Tao Deng}
\ead{tdeng@swjtu.edu.cn}

\author[label1]{Hongmei Yan \corref{cor1}}
\ead{hmyan@uestc.edu.cn}

\cortext[cor1]{Corresponding author.}
\address[label1]{MOE Key Laboratory for Neuroinformation, School of Life Science and Technology, University of Electronic Science and Technology of China, Chengdu, China}

\address[label2]{School of Information Science and Technology, Southwest Jiaotong University, Chengdu, China}


\begin{abstract}

Multi-scale detection plays an important role in object detection models. However, researchers usually feel blank on how to reasonably configure detection heads combining multi-scale features at different input resolutions. We find that there are different matching relationships between the object distribution and the detection head at different input resolutions. Based on the instructive findings, we propose a lightweight traffic object detection network based on matching between detection head and object distribution, termed as MHD-Net. It consists of three main parts. The first is the detection head and object distribution matching strategy, which guides the rational configuration of detection head, so as to leverage multi-scale features to effectively detect objects at vastly different scales. The second is the cross-scale detection head configuration guideline, which instructs to replace multiple detection heads with only two detection heads possessing of rich feature representations to achieve an excellent balance between detection accuracy, model parameters, FLOPs and detection speed. The third is the receptive field enlargement method, which combines the dilated convolution module with shallow features of backbone to further improve the detection accuracy at the cost of increasing model parameters very slightly. The proposed model achieves more competitive performance than other models on BDD100K dataset and our proposed ETFOD-v2 dataset. The code will be available.
\end{abstract}


\end{frontmatter}

\section{Introduction}
\label{sec1}

\begin{figure}[t]
	\centering
	\includegraphics[width=0.75\columnwidth]{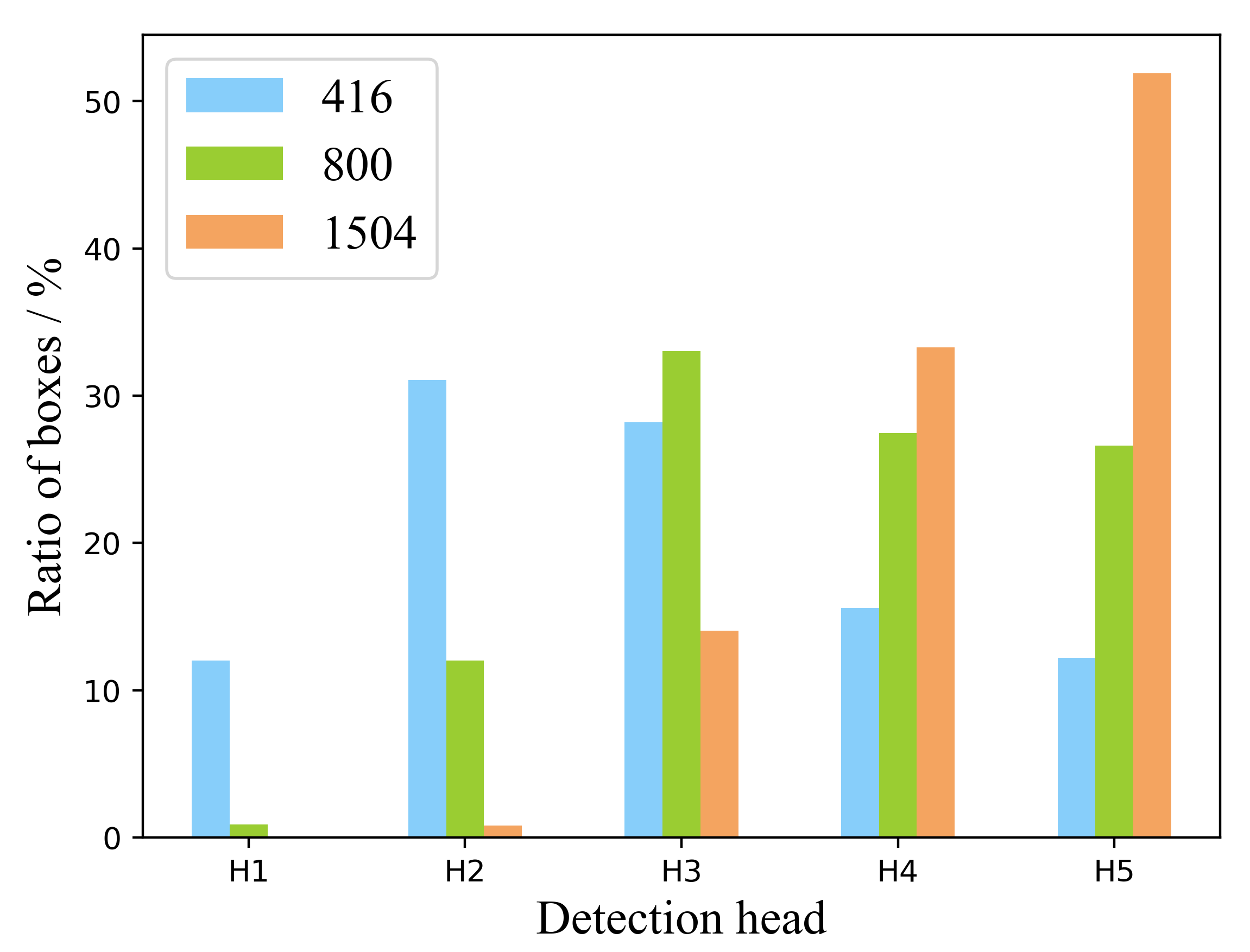} 
	\caption{The matching relationships between detection head and object distribution at different input resolution on the BDD100K training set. H1, H2, H3, H4 and H5 represent the detection heads corresponding to feature maps with the down-sampling rate of 2, 4, 8, 16 and 32, respectively.}.
	\label{fig1}
\end{figure}

As an crucial part of intelligent driving, object detection \citep{chen2021deep, liu2022image} is important for ensuring driving safety. In general, to balance the detection accuracy with FLOPs, scaling input resolution is a common method \citep{dollar2021fast,liu2022swin}. We review the details of detection models on BDD100K dataset \citep{yu2020bdd100k} and surprisingly find that different detection head can match different scale objects at different input resolutions. As shown in Figure \ref{fig1}, with low input resolution, there are a large number of objects matched with H1 and H2 detection heads. However, with the increase of input resolution, the number of objects matched with H1 and H2 detection heads is significantly decreased, while those matched with H4 and H5 detection heads is increased obviously.

This find is exciting and motivate us to rethink whether the configuration of detection heads in existing models is optimal. For example, with low input resolution, can three detection heads similar to that set by YOLOv5 \citep{c:23441} be used to achieve the best detection performance, especially for a large number of small objects? With high input resolution, will five detection heads similar to IDYOLO \citep{qin2022id} cause redundancy of detection heads and increase the difficulty of model optimization? Furthermore, for different input resolutions, how to configure appropriate detection heads to achieve better detection performance for traffic objects at very varied scales?

On the other hand, to reduce model parameters and improve detection speed, it is a common method to employ fewer detection heads. Yolov3-tiny \citep{redmon2018yolov3} and YOLOv4-tiny \citep{wang2021scaled} used H4 and H5 detection heads to detect objects. This method reduced the model parameters and improved the detection speed, but sacrificed the detection accuracy. YOLOF \citep{chen2021you} utilized single H5 detection head, which decreased model parameters and increased detection accuracy, but it is not friendly to detection of small objects. CornerNet \citep{law2018cornernet} and CenterNet \citep{duan2019centernet} achieved competitive performance with single H2 detection head, but detection performance for large objects may be inadequate. For traffic scenes with limited computing resources and different scale objects, how to configure reasonable detection heads to achieve an excellent balance between detection accuracy, model parameters, FLOPs and detection speed is a problem worthy of study.

To deal with the problems mentioned above, we first conduct a preliminary study on the influence of the configuration of the detection heads on the detection performance at different input resolutions on the BDD100K dataset. The experimental results show that the detection accuracy obtained by three detection heads (H3-5) is lower than that obtained by four detection heads (H2-5) with low resolution of 416. With high resolution of 1504, the detection accuracy achieved by H3-5 detection heads is higher than that achieved by H2-5  detection heads. We believe that this totally opposite result is mainly caused by the mismatch between the detection head and object distribution. With low input resolution, a large number of small objects match with detection heads corresponding to high resolution feature maps. In this case, only three detection heads cannot effectively detect a large number of small objects. In accordance with cognition, the addition of H2 detection head is beneficial to small object detection. With high input resolution, the object scale becomes larger as a whole, and three detection heads can almost match all objects. In this case, the use of more detection heads may cause detection head redundancy, which is not conducive to model optimization. To alleviate the detection performance degradation caused by the mismatch, based on above findings, we have made the following three contributions:

1) We propose an applicable matching strategy of detection head and object distribution. The proposed matching strategy can guide us to configure detection heads reasonably for detecting different scale objects.

2) Further, a simple and effective cross-scale detection head configuration guideline is presented. Based on this guideline, multiple detection heads can be replaced by two detection heads, which can significantly reduce model parameters and FLOPs as well as improve detection speed while maintain high detection accuracy.

3) Combining the dilated convolution module with shallow features of backbone, we construct a lightweight traffic object detection network, termed as MHD-Net. Experimental results show that our proposed model achieves more competitive performance than other models on BDD100K dataset and our proposed ETFOD-v2 dataset.

\section{Related works}\label{sec2}
\subsection{Multi-scale detection}
Multi-scale detection plays an important role in object detection models. The models represented by VIT-YOLO \citep{zhang2021vit}, FCOS \citep{tian2020fcos}, VFNet \citep{zhang2021varifocalnet} and GFL \citep{li2020generalized} utilized five different feature levels to fuse contextual information, and thus achieved impressive detection performance. Considering that the features of small objects are easy missing in the down-sampling process, Libra-RCNN \citep{pang2019libra}, OPANAS \citep{liang2021opanas}, ABFPN \citep{9720996} and GCA RCNN \citep{zhang2021global} integrated four different feature levels to construct the detection model for detection of objects under varied scales. YOLOv5 \citep{c:23441}, YOLOv7 \citep{wang2022yolov7}, CSPNet \citep{wang2020cspnet} and YOLOv4-P5 \citep{wang2021scaled} only employed the features of three scales to represent the different scale objects, and also achieve excellent performance. Different from the above multi-scale representation, YOLOv3-Tiny \citep{redmon2018yolov3}, YOLOv4-Tiny \citep{wang2021scaled} and YOEO \citep{vahl2021yoeo} integrated the features of the two scales for object detection, so as to achieve the balance between detection accuracy, model parameters and detection speed. Although these models have achieved impressive performance, they do not fully consider the matching relationships between the object distribution and the detection head at different resolutions. In our proposed method, based on the instructive findings as presented in previous section, detection heads can be simply and reasonably configured according to the matching relationship, and more appropriate and representative multi-scale information can be fully utilized to detect objects at different scales.

\subsection{Traffic object detection}
With the rise of intelligent driving, traffic object detection has received more and more attention. Yang et al. \citep{yang2020part} proposed a Part-Aware Multi-Scale Fully Convolutional Network with two detection heads for detecting pedestrian. Based on YOLOv5 model and decision tree, Aboah et al. \citep{aboah2021vision} built a traffic monitoring system to monitor abnormal traffic events, and achieved the expected results. To detect objects in railways, DFF-Net \citep{ye2020railway} with the prior object detection module and the object detection module was built and obtained better performance than other models. Lin et al. \citep{lin2020graininess} combined the attention map with multi-scale features, proposed a model that paid more attention to small-scale pedestrians, and achieved good performance on multiple datasets. Different from detecting all objects in the traffic scene, Qin et al. \citep{qin2022id} presented IDYOLO model for detecting objects within the driver's attention area, and obtained competitive results. Although the above models with multiple detection heads make use of multi-scale features to obtain good results, they do not fully consider the matching relationship between object distribution and detection heads. Based on the findings as shown in figure 1, we make full use of two cross-scale detection heads instead of multiple detection heads to construct a lightweight model, achieving an outstanding balance between detection accuracy, model parameters, FLOPs and detection speed.

\section{Proposed Methods}\label{sec3}

\begin{figure}[t]
	\centering
	\includegraphics[width=0.85\columnwidth]{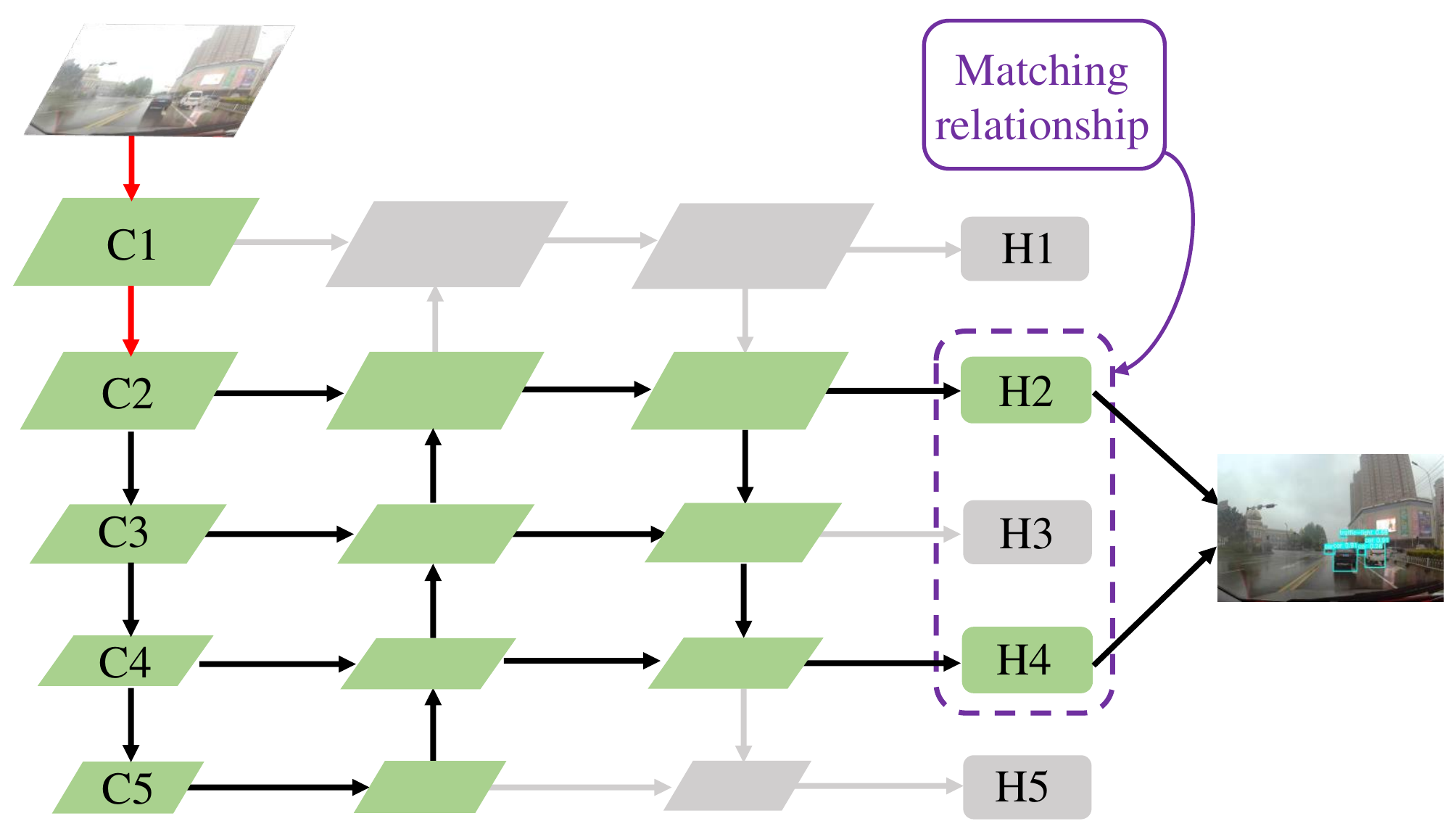} 
	\caption{The schematic diagram of MHD-Net. C1-C5 represent feature maps of backbone. The red arrow indicates the combined dilated convolution module after the first and second down-sampling layers. The matching relationship between the detection head and the object distribution guides the detection head configuration.}.
	\label{fig2}
\end{figure}

Based on the above analysis, we adopt the YOLOv5-S \citep{c:23441} model structure and replace the $6\times6$  convolution in the first down-sampling layer of backbone with $3\times3$  convolution to build a lightweight traffic object detection network, termed as MHD-Net. It consists of three main parts. The first is the detection head and object distribution matching strategy, which guides the rational configuration of detection head to effectively detect objects at vastly different scales. The second is the cross-scale detection head configuration guideline, which instructs to replace multiple detection heads with only two detection heads to obtain an excellent balance between detection accuracy, model parameters, FLOPs and detection speed. The third is the receptive field enlargement method, which combines the dilated convolution module with shallow features of backbone to further improve the detection accuracy at the cost of slightly increasing model parameters. The schematic diagram of the proposed MHD-Net is shown in Figure \ref{fig2}.

\subsection{Detection Head and Object Distribution Matching Strategy}
Although multi-scale plays an important role in object detection, fixed detection head configuration has always been a conventional model construction paradigm, such as H3-5 detection heads used by YOLOv5 \citep{c:23441}. To our surprise, we find that there are different matching relationships between the object distribution and the detection head at different input resolutions. This discovery makes us reconsider, is the existing detection head configuration paradigm the optimal choice? Furthermore, for different input resolutions, how should we configure reasonable detection heads to detect traffic objects across various scales?

With this thinking, we propose a simple and effective matching strategy of the detection head and the object distribution. We first calculate the scale range (SR) corresponding to the \textit{i-th} detection head, as shown in formula (1):
\begin{equation}
	\begin{cases} 
		\left \lceil 2^{i}\times \frac{w_{o} }{w_{in}} \right \rceil ^{2} \le SR_{i} <  \left \lceil 2^{i+1}\times \frac{w_{o} }{w_{in} } \right \rceil ^{2},  i\in \{ 1, 2,3,4 \} \\
		SR_{i} \ge \left \lceil 2^{i}\times \frac{w_{o} }{w_{in}} \right \rceil ^{2} , i\in \left \{ 5\right \}
	\end{cases}
\end{equation}
where, $w_{o}$ and $w_{in}$ represent the width of the original traffic image and the resized image, respectively. $\left \lceil \cdot  \right \rceil$  means ceiling function. Then, the matching relationship (R) between object distribution and each detection head can be calculated as follows:
\begin{equation}
	\label{Eq.f_dc} R_{i} = \frac{N\left (S_{gt}\in SR_{i} \right ) }{N_{t} } , i\in \left \{1, 2, \cdots, 5 \right \}  
\end{equation}
where, $S_{gt}$ represents the area of the object box on the training set. $N\left (\cdot \right ) $
is the number of objects that match the \textit{i-th} detection head. $N_{t}$ is the number of all objects on the training set. Given an input resolution of 800, the matching relationship between the detection head and the object distribution can be calculated with formula (1) and (2), as shown in Figure \ref{fig1}. It can be seen that a large number of objects match with H2-H5 detection heads, while H1 detection head match with less than 1 $ \% $ of objects. Naturally, we choose H2-H5 detection heads with sufficient feature representations to build the model for object detection. The detection head is configured according to the matching relationship, which can not only make each detection head utilize the divide-and-conquer strategy to detect the object of the corresponding scale from the local point of view, but also can cover the object of different scales to the maximum extent from the global point of view, resulting in better detection accuracy. Further, at different input resolutions, the proposed formula (1) and (2) can be used to calculate the matching relationship between the detection head and the object distribution, so as to guide the appropriate detection head configuration.

\subsection{Cross-scale Detection Head Configuration Guideline}
The detection head configuration based on the detection head and the object distribution matching strategy can achieve better detection accuracy. However, due to the adoption of multiple detection heads, model parameters and FLOPs will inevitably be increased. A model with less parameters and low FLOPs has high detection speed and accuracy, which is ideal for traffic object detection. YOLOF \citep{chen2021you} used only single H5 detect head to achieve competitive performance, but has poor detection result for small objects. TTFNET \citep{liu2020training} achieved comparable performance by using single H2 detection head, but there may be a risk of detection performance degradation for large objects. For the detection of traffic objects with varied scales, a simple method is to combine H2 with H5 detection head, but is this combination of fixed configuration the best choice? Furthermore, for traffic object detection at different input resolutions, how can the detection head be configured to achieve a better balance between detection accuracy, model parameters, FLOPs and detection speed?

To deal with this problem, we propose a cross-scale detection head configuration guideline based on the matching relationship between the detection head and the object distribution calculated by formula (1) and (2). Given an input resolution of 800, we have known that a large number of objects match with H2-H5 detection heads. To effectively detect small objects, we first select H2 head for detector construction. Considering the large number of objects that match H3-H5 heads, it is not enough to have just single H2 head for detecting all objects. So, we utilize cross-scale detection head configuration, that is, H4 head is selected on the basis of H2 head to jointly build the detector. The double detection heads model based on the cross-scale detection head configuration guideline can not only make full use of shallow and deep features to detect objects across various scales, but also significantly reduce model parameters and FLOPs, as well as improve detection speed. Further, based on the matching relationship, the proposed cross-scale detection head configuration guideline can guide to leverage two detection heads instead of multiple detection heads for achieving an excellent balance between detection accuracy, model parameters, FLOPs and detection speed at different input resolution.

\subsection{Receptive Field Enlargement Method}
The dilated convolution is widely used in object detection task because it can increase the receptive field with few parameters. YOLOF \citep{chen2021you}, RFB-Net \citep{liu2018receptive}, LFD-Net \citep{ye2020application} and SA-YOLOv3 \citep{tian2020sa} combined dilated convolutions with deep features to improve the detection performance, but significantly increase the model parameters. We believe that for traffic object detection, a model that can improve the detection accuracy and maintain a high detection speed while having fewer parameters is undoubtedly more promising.

\begin{figure}[!ht]
	\centering
	\includegraphics[width=0.8\columnwidth]{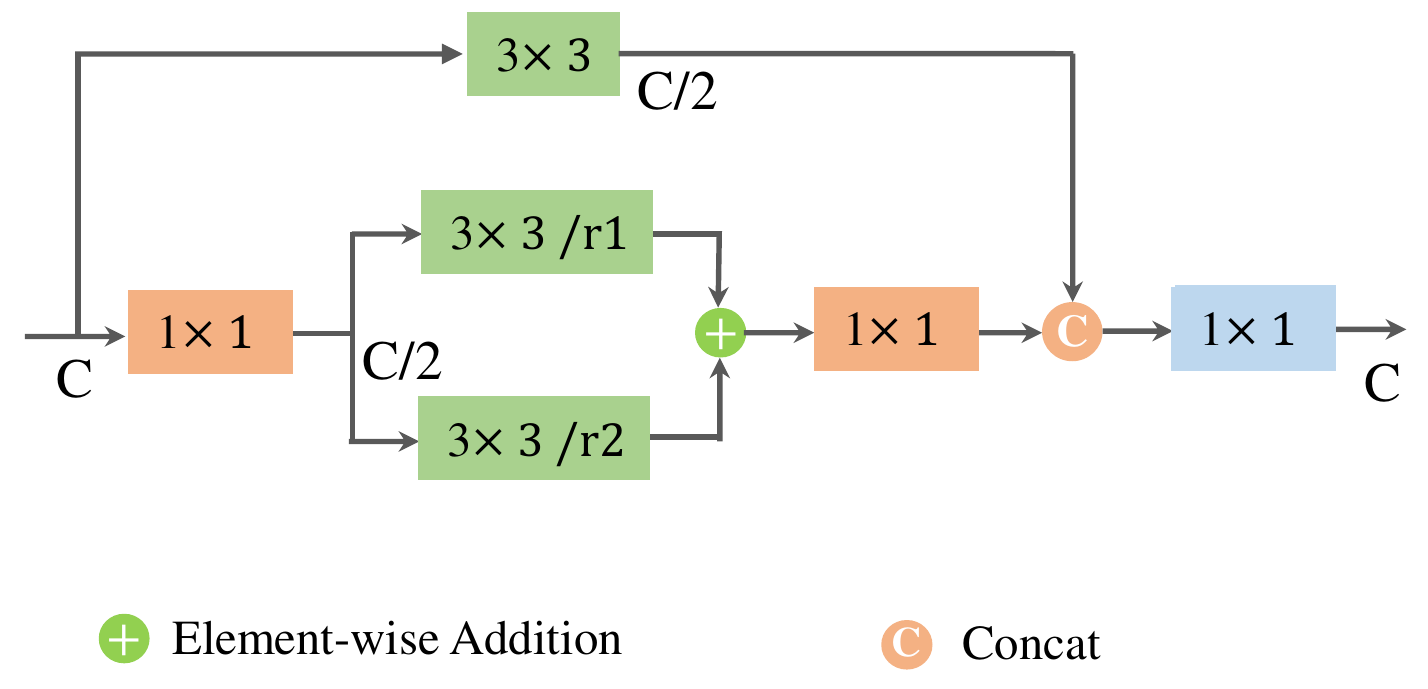} 
	\caption{Lightweight dilated convolution module. C represents channel dimension. r1 and r2 represent dilation rates with 4 and 8, respectively.}
	\label{fig3}
\end{figure}

Motivated by this vision, we propose to combine the dilated convolution module in the shallow layer of backbone to achieve excellent balance between model parameters, detection accuracy, FLOPs and detection speed. The proposed dilated convolution module is shown in figure \ref{fig3}. The core idea includes two aspects: one is to use as few convolution kernels and small convolution kernels as possible to reduce the module parameters; the other is to employ shortcut method to integrate two convolutions with different dilation rates and one standard $3\times3$ convolution to increase the model receptive field while alleviating the gridding problem \citep{wang2018understanding}. The proposed dilated convolution module is concise and lightweight. It is easy to combine with any layer of backbone.

\section{Experiments and Results}\label{sec4}
\subsection{Implementation details}
To verify the effectiveness of our proposed methods, extensive experiments are carried out on BDD100K dataset and our proposed ETFOD-v2 dataset. The proposed MHD-Net is trained by SGD optimizer with hyperparameters as same as YOLOv5-S. The initial learning rate is 0.01 and weight decay is 0.0005. Without special explanation, the batch size is set to 16 with single GeForce RTX 3090 GPU during training. Consistent with IDYOLO and LFD-Net \citep{ye2020application}, mAP, parameters, FLOPs and FPS are used as evaluation metrics.

\subsection{Influence of detection head configuration}
In this subsection, MHD-Net without the dilated convolution module is used to explore the influence of detection head configuration on detection performance at different input resolution on BDD100K dataset.


\paragraph{Input resolution of 416}

 It can be seen from Figure \ref{fig1} that a large number of objects match with five detection heads with resolution of 416. According to the proposed matching strategy of detection head and object distribution, H1-5 detection heads are set to detect objects, thus achieving the highest mAP value, as shown in Table \ref{tabe1}, which manifests that the proposed strategy is effective. In this case, the configuration of five detection heads covers different scale objects to the maximum extent, thus achieving the highest detection accuracy. 

\begin{table}[!ht]
	\centering
	
		
		\begin{tabular}{p{1.0cm}<{\centering}|p{0.7cm}<{\centering} p{0.7cm}<{\centering} p{0.7cm}<{\centering} p{0.7cm}<{\centering}|p{0.8cm}<{\centering} p{0.9cm}<{\centering} p{0.8cm}<{\centering}} 
			\toprule
			Head & mAP & $AP_{S}$ & $AP_{M}$ & $AP_{L}$ & Para. & FLOPs & FPS\\
			\midrule
			H1-5 & 48.0 & 39.0 & 62.0 & 54.8 & 7.23 & 4.54 & 79 \\
			H1,3 & 47.7 & 38.6 & 62.7 & 55.4 & 4.97 & 3.89 & 95 \\
			H1,2 & 46.9 & 38.8 & 59.5 & 50.8 & 4.86 & 3.61 & 96 \\
			H1,5 & 42.8 & 34.3 & 59.0 & 51.3 & 7.21 & 4.49 & 82 \\
			H4,5 & 32.8 & 22.8 & 59.1 & 51.9 & 6.47 & 2.67 & 130 \\
			H3-5 & 41.2 & 31.1 & 60.5 & 51.8 & 7.04 & 3.25 & 100 \\
			H2-5 & 46.3 & 37.4 & 61.1 & 53.4 & 7.19 & 3.86 & 94 \\
			\bottomrule
		\end{tabular}
		\caption{Influence of detection head configuration on detection performance with input resolution of 416.}
		\label{tabe1}
	\end{table}

Further, according to the proposed cross-scale detection head configuration guideline, we select H1 and H3 detection heads possessing of rich feature representations to detect traffic objects. As can be seen from Table \ref{tabe1}, compared with using five detection heads, mAP obtained by fully utilizing two cross-scale detection heads is only slightly declined, but model parameters are sharply reduced by more than 31 $\%$, FLOPs is also decreased by more than 14 $\%$, and FPS is improved by more than 20 $\%$. The outstanding balance between detection accuracy, model parameters, FLOPs and detection speed demonstrates the efficiency of our proposed cross-scale detection head configuration. In addition, we find that when continuous H1 and H2 detection heads are deployed, mAP value is decreased, which indicates that a large number of objects with mismatching these two detection heads may be beyond the effective perception range of detection heads, thereby reducing the detection accuracy. A similar decrease in detection accuracy occurs with H1 and H5 detection head configurations. The model constructed with H4 and H5 detection heads obtains the lowest mAP value, especially for small objects. The reason may be that the position information of a large number of small objects is easily missing in the down-sampling process. In this case, the detection head corresponding to low-resolution feature map cannot effectively represent the small object features, resulting in very poor detection results. With the addition of the detection head corresponding to high-resolution feature map, the number of small objects matching the detection head gradually increases, the representation ability of the detection head for small objects is strengthened, and the detection performance is improved.

\paragraph{Input resolution of 800}

 With input resolution of 800, the matching relationship between the object distribution and each detection head is calculated as shown in Figure \ref{fig1}. It can be found that a large number of objects match with H2-5 detection heads. Based on the proposed matching strategy of detection head and object distribution, these four detection heads are utilized to detect traffic objects. The detection results are represented in Table \ref{tabe2}. Unsurprisingly, the highest mAP value is obtained with these four detection heads configuration, which again validates the effectiveness of our proposed strategy.

\begin{table}[!ht]
	\centering
				
				\begin{tabular}{p{1.0cm}<{\centering}|p{0.7cm}<{\centering} p{0.7cm}<{\centering} p{0.7cm}<{\centering} p{0.7cm}<{\centering}|p{0.8cm}<{\centering} p{0.9cm}<{\centering} p{0.8cm}<{\centering}}
					\toprule
					Head & mAP & $AP_{S}$ & $AP_{M}$ & $AP_{L}$ & Para. & FLOPs & FPS\\
					\midrule
					H2-5 & 57.1 & 40.3 & 62.4 & 63.9 & 7.19 & 14.36 & 79 \\
					H2,4 & 56.9 & 40.1 & 64.1 & 62.0 & 5.39 & 13.08 & 84 \\
					H2,3 & 56.1 & 40.1 & 62.9 & 59.1 & 4.94 & 12.00 & 89 \\
					H2,5 & 55.5 & 39.4 & 62.0 & 62.2 & 7.17 & 14.27 & 81 \\
					H4,5 & 50.1 & 31.4 & 59.3 & 61.1 & 6.47 & 9.93 & 116 \\
					H3-5 & 55.1 & 37.4 & 62.1 & 62.3 & 7.04 & 12.10 & 87 \\
					H1-5 & 56.4 & 40.4 & 62.6 & 61.8 & 7.23 & 16.79 & 65 \\
					\bottomrule
				\end{tabular}
				\caption{Influence of detection head configuration on detection performance with input resolution of 800.}
				\label{tabe2}
			\end{table}

Further, based on the proposed cross-scale detection head configuration guideline, H2 and H4 detection heads are employed to obtain an excellent balance between detection accuracy, model parameters, FLOPs and detection speed. Consistent with the detection performance with input resolution of 416, the configuration of continuous H2 and H3 detection heads reduces detection accuracy. The mAP obtained with H4 and H5 detection heads is the lowest. With the addition of the detection head corresponding to high-resolution feature maps, the detection performance of small objects is improved. However, more detection heads is not always better. The detection performance obtained by using five detection heads is not as good as that obtained by using four detection heads. The detection accuracy and speed are reduced while the parameters and FLOPs are increased. The reason may be that a large number of objects match with H2-5 detection heads, while less than 1 $\%$ of the objects match with the H1 detection head. In this case, adding H1 detection head will lead to redundancy of detection head, which is not conducive to model optimization. 

\paragraph{Input resolution of 1504}
As can be seen from Figure \ref{fig1}, with input resolution of 1504, a large number of objects match with H3-5 detection heads, while few objects match with H1 and H2 detection heads, accounting for less than 1 $\%$. Based on the proposed matching strategy of detection head and object distribution, H3-5 detection heads are employed to detect objects and obtain the highest mAP value, as listed in Table \ref{tabe3}. Further, based on the proposed cross-scale detection head configuration guideline, H3 and H5 detection heads are leveraged to obtain the same mAP value as with H3-5 detection heads. Although the decrease of model parameters and the increase of FPS based on the proposed cross-scale detection heads H3,5 configuration are just slightly better than those using H3-5 detection heads, the guiding significance of the proposed cross-scale detection head configuration guideline is still effective. Consistent with the detection performance with input resolution of 416 and 800, the mAP obtained with H4 and H5 detection heads is the lowest. With the addition of the detection head corresponding to high-resolution feature map, the detection performance of small objects is improved. Similarly, when the detection head that do not match with a lot of objects is continued to be added, for example H2 head, the detection accuracy and speed are decreased, also the model parameters and FLOPs are increased.

\begin{table}[t]
	\centering
		
		\begin{tabular}{p{1.0cm}<{\centering}|p{0.7cm}<{\centering} p{0.7cm}<{\centering} p{0.7cm}<{\centering} p{0.7cm}<{\centering}|p{0.8cm}<{\centering} p{0.9cm}<{\centering} p{0.8cm}<{\centering}}
			
			\toprule
			Head & mAP & $AP_{S}$ & $AP_{M}$ & $AP_{L}$ & Para. & FLOPs & FPS\\
			\midrule
			H3-5 & 60.5&33.2&57.4&70.0& 7.04 & 42.78 & 75 \\
			H3,5 & 60.5&33.1&56.4&70.6& 7.03 & 42.68 & 77 \\
			H3,4 & 59.8&33.6&56.7&68.3& 5.25 & 38.80 & 83 \\
			H4,5 & 58.9&30.1&56.1&70.1& 6.47 & 35.10 & 100 \\
			H2-5 & 59.9&33.1&56.3&69.2& 7.19 & 50.40 & 65 \\		
			\bottomrule
		\end{tabular}
		\caption{Influence of detection head configuration on detection performance with input resolution of 1504.}
		\label{tabe3}
	\end{table}

In general, the detection head configuration has a great impact on the detection performance at different input resolutions. Based on our proposed detection head and object distribution matching strategy, an applicable model can be constructed to achieve the highest detection accuracy than other detection head configuration. Based on our proposed cross-scale detection head configuration guideline, only two detection heads can be leveraged instead of multiple detection heads to achieve splendid balance between detection accuracy, model parameters, FLOPs and detection speed.

\begin{table}[t]
	\centering
		
		\begin{tabular}{p{2.2cm}<{\centering}|p{0.55cm}<{\centering} p{0.5cm}<{\centering} p{0.5cm}<{\centering} p{0.5cm}<{\centering} p{0.5cm}<{\centering} p{0.5cm}<{\centering} p{0.5cm}<{\centering} p{0.5cm}<{\centering} p{0.5cm}<{\centering} p{0.7cm}<{\centering} p{0.75cm}<{\centering} p{0.85cm}<{\centering} p{0.6cm}<{\centering}}
			
				
				\toprule
				Methods & mAP & Car & Bus & Ped. & Bic. & Tru. & Mot. & Rid. & Sign & Light & Para. & FLOPs & FPS\\ 
				\midrule
				
				YOLOv5-X  & 51.0 & 74.8 & 59.7 & 56.1 & 44.9 & 61.9 & 47.4 & 43.1 & 62.4 & 59.5 & 86.28 & 43.27 & 56  \\
				YOLOv5-M  & 46.3 & 71.8 & 55.4 & 50.9 & 38.7 & 57.7 & 39.1 & 37.9 & 57.0 & 54.5 & 20.91 & 10.17 & 79  \\
				YOLOv4-P5  & 51.0 & 76.9 & 60.3 & 56.8 & 44.5 & 61.9 & 43.1 & 41.5 & 64.3 & 60.3 & 70.33 & 35.10 & 47  \\
				IDYOLO & 48.5 & 78.1 & 57.5 & 57.2 & 39.8 & 58.3 & 38.8 & 37.8 & 61.7 & 55.4 & 34.03 & 42.95  & 78 \\
				Sparse-Rcnn & 42.8 & 69.6 & 49.5 & 46.2 & 32.1 & 52.1 & 35.8 & 34.9 & 55.8 & 51.9 & 124.95 & 44.66 & 30  \\
				PVT-tiny & 45.2 & 70.6 & 54.5 & 53.4 & 44.8 & 53.5 & 39.8 & 42.3 & 52.7 & 40.6 & 21.51 & 77.33 & 39  \\
				PVTv2-B0 & 46.8 & 72.1 & 55.4 & 55.3 & 45.2 & 56.6 & 42.6 & 42.0 & 54.7 & 43.8 & 11.49 & 61.78 & 37  \\
				
				YOLOX-S & 40.7 & 69.9 & 48.8 & 44.5 & 29.8 & 51.6 & 32.7 & 31.1 & 49.2 & 49.0 & 8.94 & 5.63 & 79 \\
				YOLOX-tiny & 38.8 & 69.2 & 44.6 & 43.9 & 27.0 & 49.0 & 27.8 & 28.8 & 48.9 & 48.6 & 5.04 & 3.20 & 84 \\
				YOLOv4-tiny  & 32.1 & 57.5 & 52.2 & 25.1 & 30.4 & 51.1 & 29.6 & 30.8 & 27.7 & 16.7 & 5.89 & 3.42 & 168 \\
				YOLOv5-S & 41.0 & 68.1 & 50.9 & 44.1 & 31.5 & 53.6 & 33.9 & 31.5 & 50.0 & 46.7 & 7.05 & 3.38 & 98 \\
				YOLOv7-tiny & 40.9 & 71.0 & 52.8 & 45.2 & 29.9 & 52.6 & 32.5 & 31.3 & 47.8 & 45.7 & 6.04 & 2.80 & 103 \\ 
				\midrule
				MHD-Net & 50.3 & 78.2 & 57.9 & 58.3 & 41.3 & 60.3 & 41.9 & 42.4 & 61.2 & 61.8 & 5.03 & 4.88 & 84 \\
				\bottomrule		
			\end{tabular}
			\caption{Comparison of detection results of state-of-the-art models (YOLOv5-X, YOLOv5-M and YOLOv5-S \citep{c:23441}; YOLOv4-P5 and YOLOv4-tiny \citep{wang2021scaled}; IDYOLO \citep{qin2022id}; Sparse-Rcnn \citep{sun2021sparse}; PVT-tiny \citep{wang2021pyramid}; PVTv2-B0 \citep{wang2022pvt}; YOLOX-S and YOLOX-tiny \citep{ge2021yolox}; YOLOv7-tiny \citep{wang2022yolov7} ) on the validation set. The $AP_{train}$ obtained by most of the models is 0, so the $AP_{train}$ is not listed in the table. PVT-tiny and PVTv2-B0 are RetinaNet-PVT-tiny and RetinaNet-PVTv2-B0, respectively.}
			\label{tabe4}
		\end{table}

\subsection{Comparison with State-of-the-Art}
In this subsection, the proposed MHD-Net is compared with some state-of-the-art models on the BDD100K dataset. For a fair comparison, all models use an input resolution of 416, except for the PVTv2-B0 and PVT-tiny models with input resolution of 896.

The detection results are presented in Table \ref{tabe4}. It can be seen that YOLOv4-tiny configured with H4 and H5 detection heads achieves the lowest mAP value. The detection result is consistent with those presented in the previous subsection. YOLOv5-X and YOLOv4-P5 achieve the highest mAP, but they have large parameters and high FLOPs, as well as low FPS. Based on the proposed cross-scale detection head configuration guideline, H1 and H3 detection heads are leveraged to build a lightweight model. On this basis, we combine the proposed dilated convolution module to build a new detection model MHD-Net. Compared with YOLOv5-X, the mAP value of MHD-Net only decreases by 0.7, but the model parameters are sharply reduced by more than 94 $\%$, FLOPs is significantly reduced by more than 88 $\%$, and the detection speed is improved by 50 $\%$. The proposed MHD-Net achieve splendid balance between detection accuracy, parameters, FLOPs and detection speed. In addition, we observe that except for the  $AP_{train}$ obtained by IDYOLO, YOLOv4-P5 and PVT-tiny is 0.1, the $AP_{train}$ obtained by other models is all 0, which hint that the detection of severely unbalanced objects is still a problem worth our efforts to solve.

\subsection{Influence of dilated convolution}
In this subsection, a lightweight model configurated with H1 and H3 detection heads is used as the baseline, and the proposed dilated convolution module combine with different down-sampling layers of backbone of the baseline to explore the impact of dilated convolutions on detection performance. It can be found from Table \ref{tabe5} that mAP value is increased by 1.6 combined the dilated convolution module with the first down-sampling layer, while the model parameters are only increased by 0.01M. Combining the dilated convolution module with the first and second down-sampling layers improves mAP values by 2.6 and model parameters by only 0.06M. When the dilated convolution module is further integrated in deep layers of backbone, mAP value increases slightly, but the model parameters and FLOPs increase significantly. The reason may be that our proposed dilated convolution module makes use of large dilation rates to generate a large receptive field, which has obvious gain for shallow features with high resolution, but the gain will be weak for deep features with low resolution. To better balance the detection accuracy, parameters, FLOPs and detection speed, we combine the dilated convolution module with two shallow down-sampling layers to build a lightweight detection model for traffic object detection.

\begin{table}[t]
	\centering
		\begin{tabular}{c|cccc|ccc}
			\toprule
			Layer & \multicolumn{1}{p{1.5em}}{mAP} & \multicolumn{1}{p{1.5em}}{$AP_{S}$} & \multicolumn{1}{p{1.5em}}{$AP_{M}$} & \multicolumn{1}{p{1.5em}|}{$AP_{L}$} & \multicolumn{1}{p{1.5em}}{Para.} & \multicolumn{1}{p{3.0em}}{FLOPs} &\multicolumn{1}{p{1.5em}}{FPS} \\
			\midrule		
			H1,3 & 47.7 & 38.6 & 62.7 & 55.4 & 4.97 & 3.89 & 95 \\
			L1 & 49.3 & 40.7 & 62.9 & 53.7 & 4.98 & 4.39 & 89 \\
			L1,2 & 50.3 & 41.6 & 63.0 & 56.8 & 5.03 & 4.88 & 84 \\
			L1-3 & 50.4 & 41.7 & 63.6 & 53.9 & 5.21 & 5.36 & 76 \\
			L1-4 & 50.5 & 41.8 & 64.5 & 54.9 & 5.91 & 5.84 & 70 \\
			L1-5 & 50.6 & 41.7 & 64.7 & 56.2 & 8.73 & 6.31 & 65 \\
			\bottomrule
		\end{tabular}
		\caption{Influence of dilated convolution on detection performance with resolution of 416. H1,3 represents the baseline.}
		\label{tabe5}
	\end{table}

\subsection{Detection on ETFOD-v2 dataset}
To verify the generalization performance of the proposed detection head configuration method and MHD-Net, a series of experiments are conducted on our proposed ETFOD-v2 dataset. The ETFOD-v2 dataset is an extended version of the ETFOD dataset \citep{shi4111797traffic} with the addition of 2,750 traffic images cotaining 5 object categories (traffic sign, person, car, motorcycle and traffic light). The ETFOD-v2 dataset focuses on detecting traffic objects within the driver's fixation area \citep{qin2022id}. In the experiment, 9774 images, 1455 images and 2716 images are used for training, validating and testing, respectively.

\begin{figure}[t]
	\centering
	\includegraphics[width=0.75\columnwidth]{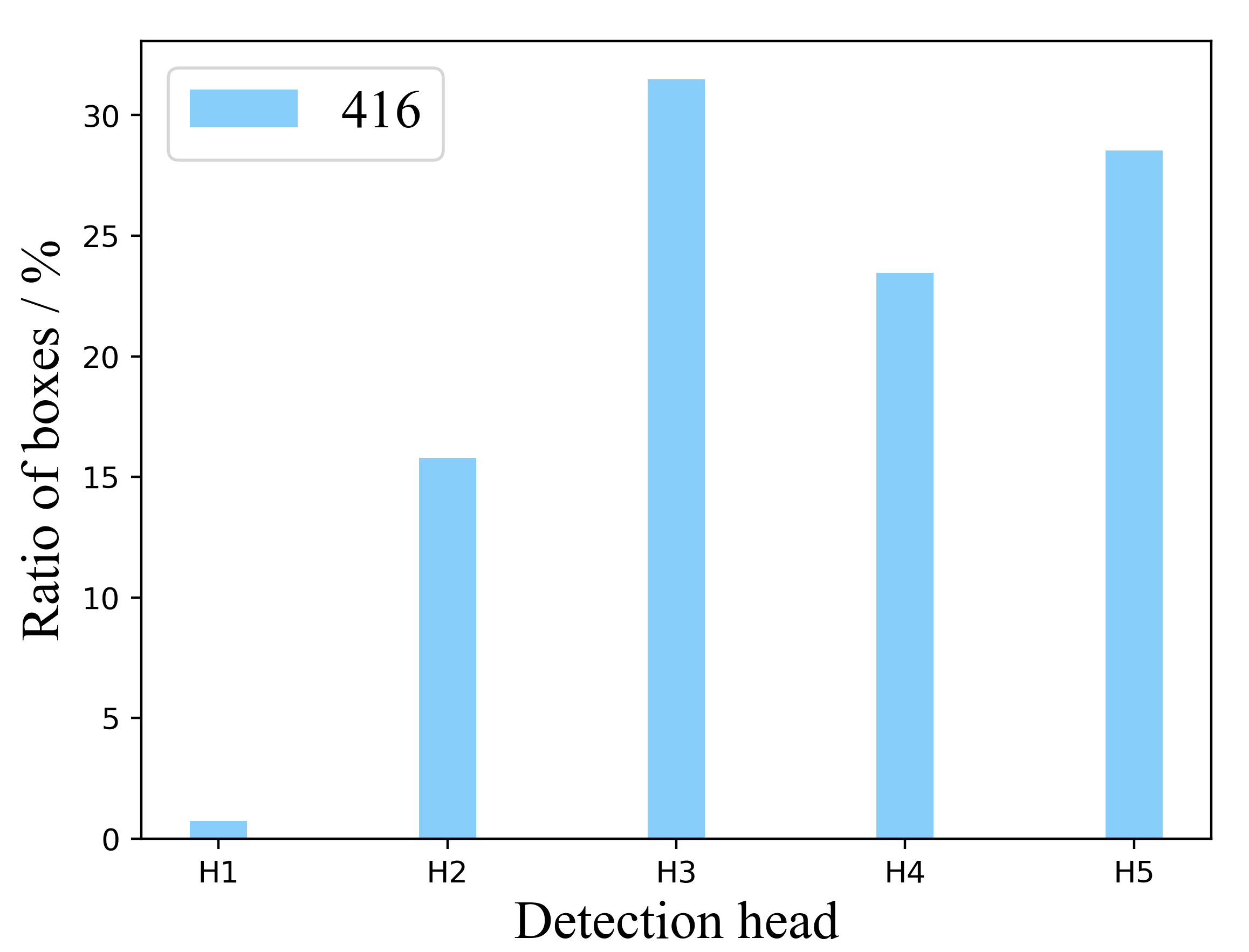} 
	\caption{The matching relationship between detection head and object distribution with resolution of 416.}.
	\label{fig4}
\end{figure}

The matching relationship between the object distribution and the detection head is calculated on the training set, as shown in Figure \ref{fig4}. It can be seen that with input resolution of 416, a large number of objects match with H2-5 detection heads, while few objects match with H1 detection head, less than 1 $\%$. Based on the proposed matching strategy of detection head and object distribution, H2-5 detection heads are configurated to achieve the highest mAP value, as shown in Table 6, which verify the generalization performance of our proposed strategy. Based on the proposed cross-scale detection head configuration guideline, H2 and H4 detection heads are leveraged to build a lightweight model for obtaining an excellent balance between detection accuracy, model parameters, FLOPs and detection speed. In addition, the experimental results again show that the detection performance obtained by other detection head configurations is inadequate. The mAP value obtained by using H4 and H5 detection heads is the lowest, and mAP value becomes higher with the addition of the detection head corresponding to high-resolution feature map. However, more detection heads are not always better. When the detection head that do not match with a lot of objects is continued to be added, the detection performance is also degraded.

	\begin{table}[h]
	\centering
		
			\begin{tabular}{c|cccc|ccc} 
				\toprule
				Head & \multicolumn{1}{p{1.6em}}{mAP} & \multicolumn{1}{p{1.6em}}{$AP_{S}$} & \multicolumn{1}{p{1.6em}}{$AP_{M}$} & \multicolumn{1}{p{1.6em}|}{$AP_{L}$} & \multicolumn{1}{p{1.8em}}{Para.} & \multicolumn{1}{p{3.0em}}{FLOPs} &\multicolumn{1}{p{1.6em}}{FPS} \\ 
				\midrule
				H2-5 & 68.5 & 67.0 & 60.3 & 91.5 & 7.18 & 3.86 & 94 \\ 
				H2,4 & 68.3 & 66.2 & 59.6 & 91.0 & 5.38 & 3.55 & 99 \\ 
				H2,3 & 67.5 & 65.9 & 59.0 & 89.1 & 4.93 & 3.25 & 103 \\ 
				H4,5 & 61.4 & 57.4 & 57.9 & 90.1 & 6.46 & 2.67 & 130 \\ 
				H3-5 & 65.7 & 63.1 & 60.2  & 91.6  & 7.03 & 3.24 & 100 \\ 
				H1-5 & 67.4 & 64.9  & 60.1  & 90.4  & 7.21 & 4.49  & 80 \\ 
				\bottomrule
			\end{tabular}
			\caption{Influence of detection head configuration on detection performance with input resolution of 416.}
			\label{table6}
		\end{table}

Further, we leverage H2 and H4 detection heads and combine the proposed dilated convolution module to construct a lightweight detection model, MHD-Net. The proposed MHD-Net is compared with some state-of-the-art models on the ETFOD-v2 dataset. As can be seen from Table \ref{tabe7}, YOLOv4-tiny achieves the lowest mAP value. Although PVTv2-B0 \citep{wang2022pvt} achieves the highest mAP value, its parameters, FLOPs and detection speed are nearly 2.28 times, 12.65 times and 0.44 times of our proposed MHD-Net, respectively. The excellent balance between detection accuracy, model parameters, FLOPs and detection speed proves the generalization performance of our proposed model and its potential application prospects.

\begin{table}[t]
	\centering
		\begin{tabular}{c|c|ccccc}
			\toprule
			Methods & \multicolumn{1}{p{1.6em}|}{mAP} & \multicolumn{1}{p{2.0em}}{Sign} & \multicolumn{1}{p{2.0em}}{Per.} & \multicolumn{1}{p{1.6em}}{Car} & \multicolumn{1}{p{2.0em}}{Moto.} & \multicolumn{1}{p{1.6em}}{Light} \\
			\midrule
			
			YOLOv5-X & 70.6 & 72.8 & 54.1 & 93.2 & 68.9 & 64.1 \\
			YOLOv5-M & 67.9 & 69.8 & 52.5 & 92.5 & 66.2 & 58.4 \\
			IDYOLO & 64.5 & 67.8 & 44.6 & 91.9 & 62.3 & 56.1 \\
			PVTv2-B0 & 71.6 & 73.7 & 57.4 & 92.5 & 74.9 & 59.5 \\
			YOLOX-S & 64.9 & 67.3 & 44.3 & 91.1 & 68.1 & 53.8 \\
			YOLOv4-tiny & 52.5 & 62.1 & 39.5 & 81.8 & 51.8 & 27.1 \\
			YOLOv5-S & 65.6 & 69.6 & 50.5 & 91.8 & 55.3 & 60.8 \\
			YOLOv7-tiny & 66.6 & 69.2 & 50.8 & 90.3 & 68.6 & 54.2 \\
			\midrule
			MHD-Net & 71.1 & 73.1 & 54.0 & 93.5 & 70.1 & 65.0 \\
			\bottomrule
		\end{tabular}
		\caption{ Comparison of detection results of state-of-the-art models on the testing set. PVTv2-B0 is RetinaNet-PVTv2-B0 with input resolution of 896.}
		\label{tabe7}
	\end{table}

\section{Conclusion}\label{sec5}
Different from the fixed or blind detection head configuration paradigm in the existing models, based on the instructive finding that there are different matching relationships between the detection head and the object distribution at different input resolutions, we first propose an applicable matching strategy of the detection head and the object distribution to guide the reasonable configuration of detection heads for achieving the highest detection accuracy. Further, a simple and effective cross-scale detection head configuration guideline is presented to guide configurating only two detection heads instead of multiple detection heads to achieve an efficient balance between detection accuracy, model parameters, FLOPs and detection speed. Based on the lightweight detection heads configuration proposed, the dilated convolution module is combined with the shallow features of backbone to build a new lightweight model, which achieves more competitive performance compared with other models on BDD100K dataset and ETFOD-v2 dataset. We hope that our work will provide a valuable reference for future research.


\bibliography{mybib.bib}

\begin{thebibliography}{37}
\expandafter\ifx\csname natexlab\endcsname\relax\def\natexlab#1{#1}\fi
\providecommand{\url}[1]{\texttt{#1}}
\providecommand{\href}[2]{#2}
\providecommand{\path}[1]{#1}
\providecommand{\DOIprefix}{doi:}
\providecommand{\ArXivprefix}{arXiv:}
\providecommand{\URLprefix}{URL: }
\providecommand{\Pubmedprefix}{pmid:}
\providecommand{\doi}[1]{\href{http://dx.doi.org/#1}{\path{#1}}}
\providecommand{\Pubmed}[1]{\href{pmid:#1}{\path{#1}}}
\providecommand{\bibinfo}[2]{#2}
\ifx\xfnm\relax \def\xfnm[#1]{\unskip,\space#1}\fi
\bibitem[{Aboah(2021)}]{aboah2021vision}
\bibinfo{author}{Aboah, A.} (\bibinfo{year}{2021}).
\newblock \bibinfo{title}{A vision-based system for traffic anomaly detection
  using deep learning and decision trees}.
\newblock In {\it \bibinfo{booktitle}{Proceedings of the IEEE/CVF Conference on
  Computer Vision and Pattern Recognition}\/} (pp.
  \bibinfo{pages}{4207--4212}).
\bibitem[{Chen et~al.(2021{\natexlab{a}})Chen, Lin, Lu, Cao, Wu, Guo, Liu \&
  Wang}]{chen2021deep}
\bibinfo{author}{Chen, L.}, \bibinfo{author}{Lin, S.}, \bibinfo{author}{Lu,
  X.}, \bibinfo{author}{Cao, D.}, \bibinfo{author}{Wu, H.},
  \bibinfo{author}{Guo, C.}, \bibinfo{author}{Liu, C.}, \&
  \bibinfo{author}{Wang, F.-Y.} (\bibinfo{year}{2021}{\natexlab{a}}).
\newblock \bibinfo{title}{Deep neural network based vehicle and pedestrian
  detection for autonomous driving: A survey}.
\newblock {\it \bibinfo{journal}{IEEE Transactions on Intelligent
  Transportation Systems}\/},  {\it \bibinfo{volume}{22}\/},
  \bibinfo{pages}{3234--3246}.
\bibitem[{Chen et~al.(2021{\natexlab{b}})Chen, Wang, Yang, Zhang, Cheng \&
  Sun}]{chen2021you}
\bibinfo{author}{Chen, Q.}, \bibinfo{author}{Wang, Y.}, \bibinfo{author}{Yang,
  T.}, \bibinfo{author}{Zhang, X.}, \bibinfo{author}{Cheng, J.}, \&
  \bibinfo{author}{Sun, J.} (\bibinfo{year}{2021}{\natexlab{b}}).
\newblock \bibinfo{title}{You only look one-level feature}.
\newblock In {\it \bibinfo{booktitle}{Proceedings of the IEEE/CVF conference on
  computer vision and pattern recognition}\/} (pp.
  \bibinfo{pages}{13039--13048}).
\bibitem[{Doll{\'a}r et~al.(2021)Doll{\'a}r, Singh \&
  Girshick}]{dollar2021fast}
\bibinfo{author}{Doll{\'a}r, P.}, \bibinfo{author}{Singh, M.}, \&
  \bibinfo{author}{Girshick, R.} (\bibinfo{year}{2021}).
\newblock \bibinfo{title}{Fast and accurate model scaling}.
\newblock In {\it \bibinfo{booktitle}{Proceedings of the IEEE/CVF Conference on
  Computer Vision and Pattern Recognition}\/} (pp. \bibinfo{pages}{924--932}).
\bibitem[{Duan et~al.(2019)Duan, Bai, Xie, Qi, Huang \&
  Tian}]{duan2019centernet}
\bibinfo{author}{Duan, K.}, \bibinfo{author}{Bai, S.}, \bibinfo{author}{Xie,
  L.}, \bibinfo{author}{Qi, H.}, \bibinfo{author}{Huang, Q.}, \&
  \bibinfo{author}{Tian, Q.} (\bibinfo{year}{2019}).
\newblock \bibinfo{title}{Centernet: Keypoint triplets for object detection}.
\newblock In {\it \bibinfo{booktitle}{Proceedings of the IEEE/CVF international
  conference on computer vision}\/} (pp. \bibinfo{pages}{6569--6578}).
\bibitem[{Ge et~al.(2021)Ge, Liu, Wang, Li \& Sun}]{ge2021yolox}
\bibinfo{author}{Ge, Z.}, \bibinfo{author}{Liu, S.}, \bibinfo{author}{Wang,
  F.}, \bibinfo{author}{Li, Z.}, \& \bibinfo{author}{Sun, J.}
  (\bibinfo{year}{2021}).
\newblock \bibinfo{title}{Yolox: Exceeding yolo series in 2021}.
\newblock {\it \bibinfo{journal}{arXiv preprint arXiv:2107.08430}\/}, .
\bibitem[{Jocher et~al.(2022)Jocher, Chaurasia, Stoken, Borovec \&
  Kwon}]{c:23441}
\bibinfo{author}{Jocher, G.}, \bibinfo{author}{Chaurasia, A.},
  \bibinfo{author}{Stoken, A.}, \bibinfo{author}{Borovec, J.}, \&
  \bibinfo{author}{Kwon, Y.} (\bibinfo{year}{2022}).
\newblock \bibinfo{title}{{ultralytics/yolov5:v6.1}}.
\newblock \bibinfo{howpublished}{\url{https://doi.org/10.5281/zenodo.6222936}}.
\bibitem[{Law \& Deng(2018)}]{law2018cornernet}
\bibinfo{author}{Law, H.}, \& \bibinfo{author}{Deng, J.}
  (\bibinfo{year}{2018}).
\newblock \bibinfo{title}{Cornernet: Detecting objects as paired keypoints}.
\newblock In {\it \bibinfo{booktitle}{Proceedings of the European conference on
  computer vision (ECCV)}\/} (pp. \bibinfo{pages}{734--750}).
\bibitem[{Li et~al.(2020)Li, Wang, Wu, Chen, Hu, Li, Tang \&
  Yang}]{li2020generalized}
\bibinfo{author}{Li, X.}, \bibinfo{author}{Wang, W.}, \bibinfo{author}{Wu, L.},
  \bibinfo{author}{Chen, S.}, \bibinfo{author}{Hu, X.}, \bibinfo{author}{Li,
  J.}, \bibinfo{author}{Tang, J.}, \& \bibinfo{author}{Yang, J.}
  (\bibinfo{year}{2020}).
\newblock \bibinfo{title}{Generalized focal loss: Learning qualified and
  distributed bounding boxes for dense object detection}.
\newblock {\it \bibinfo{journal}{Advances in Neural Information Processing
  Systems}\/},  {\it \bibinfo{volume}{33}\/}, \bibinfo{pages}{21002--21012}.
\bibitem[{Liang et~al.(2021)Liang, Wang, Tang, Hu \& Ling}]{liang2021opanas}
\bibinfo{author}{Liang, T.}, \bibinfo{author}{Wang, Y.}, \bibinfo{author}{Tang,
  Z.}, \bibinfo{author}{Hu, G.}, \& \bibinfo{author}{Ling, H.}
  (\bibinfo{year}{2021}).
\newblock \bibinfo{title}{Opanas: One-shot path aggregation network
  architecture search for object detection}.
\newblock In {\it \bibinfo{booktitle}{Proceedings of the IEEE/CVF Conference on
  Computer Vision and Pattern Recognition}\/} (pp.
  \bibinfo{pages}{10195--10203}).
\bibitem[{Lin et~al.(2020)Lin, Lu, Wang \& Zhou}]{lin2020graininess}
\bibinfo{author}{Lin, C.}, \bibinfo{author}{Lu, J.}, \bibinfo{author}{Wang,
  G.}, \& \bibinfo{author}{Zhou, J.} (\bibinfo{year}{2020}).
\newblock \bibinfo{title}{Graininess-aware deep feature learning for robust
  pedestrian detection}.
\newblock {\it \bibinfo{journal}{IEEE Transactions on Image Processing}\/},
  {\it \bibinfo{volume}{29}\/}, \bibinfo{pages}{3820--3834}.
\bibitem[{Liu et~al.(2018)Liu, Huang et~al.}]{liu2018receptive}
\bibinfo{author}{Liu, S.}, \bibinfo{author}{Huang, D.} et~al.
  (\bibinfo{year}{2018}).
\newblock \bibinfo{title}{Receptive field block net for accurate and fast
  object detection}.
\newblock In {\it \bibinfo{booktitle}{Proceedings of the European conference on
  computer vision (ECCV)}\/} (pp. \bibinfo{pages}{385--400}).
\bibitem[{Liu et~al.(2022{\natexlab{a}})Liu, Ren, Yu, Guo, Zhu \&
  Zhang}]{liu2022image}
\bibinfo{author}{Liu, W.}, \bibinfo{author}{Ren, G.}, \bibinfo{author}{Yu, R.},
  \bibinfo{author}{Guo, S.}, \bibinfo{author}{Zhu, J.}, \&
  \bibinfo{author}{Zhang, L.} (\bibinfo{year}{2022}{\natexlab{a}}).
\newblock \bibinfo{title}{Image-adaptive yolo for object detection in adverse
  weather conditions}.
\newblock In {\it \bibinfo{booktitle}{Proceedings of the AAAI Conference on
  Artificial Intelligence}\/} (pp. \bibinfo{pages}{1792--1800}).
\newblock volume~\bibinfo{volume}{36}.
\bibitem[{Liu et~al.(2022{\natexlab{b}})Liu, Hu, Lin, Yao, Xie, Wei, Ning, Cao,
  Zhang, Dong et~al.}]{liu2022swin}
\bibinfo{author}{Liu, Z.}, \bibinfo{author}{Hu, H.}, \bibinfo{author}{Lin, Y.},
  \bibinfo{author}{Yao, Z.}, \bibinfo{author}{Xie, Z.}, \bibinfo{author}{Wei,
  Y.}, \bibinfo{author}{Ning, J.}, \bibinfo{author}{Cao, Y.},
  \bibinfo{author}{Zhang, Z.}, \bibinfo{author}{Dong, L.} et~al.
  (\bibinfo{year}{2022}{\natexlab{b}}).
\newblock \bibinfo{title}{Swin transformer v2: Scaling up capacity and
  resolution}.
\newblock In {\it \bibinfo{booktitle}{Proceedings of the IEEE/CVF Conference on
  Computer Vision and Pattern Recognition}\/} (pp.
  \bibinfo{pages}{12009--12019}).
\bibitem[{Liu et~al.(2020)Liu, Zheng, Xu, Yang, Liu \& Cai}]{liu2020training}
\bibinfo{author}{Liu, Z.}, \bibinfo{author}{Zheng, T.}, \bibinfo{author}{Xu,
  G.}, \bibinfo{author}{Yang, Z.}, \bibinfo{author}{Liu, H.}, \&
  \bibinfo{author}{Cai, D.} (\bibinfo{year}{2020}).
\newblock \bibinfo{title}{Training-time-friendly network for real-time object
  detection}.
\newblock In {\it \bibinfo{booktitle}{proceedings of the AAAI conference on
  artificial intelligence}\/} (pp. \bibinfo{pages}{11685--11692}).
\newblock volume~\bibinfo{volume}{34}.
\bibitem[{Pang et~al.(2019)Pang, Chen, Shi, Feng, Ouyang \&
  Lin}]{pang2019libra}
\bibinfo{author}{Pang, J.}, \bibinfo{author}{Chen, K.}, \bibinfo{author}{Shi,
  J.}, \bibinfo{author}{Feng, H.}, \bibinfo{author}{Ouyang, W.}, \&
  \bibinfo{author}{Lin, D.} (\bibinfo{year}{2019}).
\newblock \bibinfo{title}{Libra r-cnn: Towards balanced learning for object
  detection}.
\newblock In {\it \bibinfo{booktitle}{Proceedings of the IEEE/CVF conference on
  computer vision and pattern recognition}\/} (pp. \bibinfo{pages}{821--830}).
\bibitem[{Qin et~al.(2022)Qin, Shi, He, Zhang, Zhang, Li, Deng \&
  Yan}]{qin2022id}
\bibinfo{author}{Qin, L.}, \bibinfo{author}{Shi, Y.}, \bibinfo{author}{He, Y.},
  \bibinfo{author}{Zhang, J.}, \bibinfo{author}{Zhang, X.},
  \bibinfo{author}{Li, Y.}, \bibinfo{author}{Deng, T.}, \&
  \bibinfo{author}{Yan, H.} (\bibinfo{year}{2022}).
\newblock \bibinfo{title}{Id-yolo: Real-time salient object detection based on
  the driver's fixation region}.
\newblock {\it \bibinfo{journal}{IEEE Transactions on Intelligent
  Transportation Systems}\/}, .
\bibitem[{Redmon \& Farhadi(2018)}]{redmon2018yolov3}
\bibinfo{author}{Redmon, J.}, \& \bibinfo{author}{Farhadi, A.}
  (\bibinfo{year}{2018}).
\newblock \bibinfo{title}{Yolov3: An incremental improvement}.
\newblock {\it \bibinfo{journal}{arXiv preprint arXiv:1804.02767}\/}, .
\bibitem[{Shi et~al.(2022)Shi, Zhao, Wu, Wu \& Yan}]{shi4111797traffic}
\bibinfo{author}{Shi, Y.}, \bibinfo{author}{Zhao, S.}, \bibinfo{author}{Wu,
  J.}, \bibinfo{author}{Wu, Z.}, \& \bibinfo{author}{Yan, H.}
  (\bibinfo{year}{2022}).
\newblock \bibinfo{title}{Traffic fixated object detection guided by region and
  category}.
\newblock {\it \bibinfo{journal}{Available at SSRN 4111797}\/}, .
\bibitem[{Sun et~al.(2021)Sun, Zhang, Jiang, Kong, Xu, Zhan, Tomizuka, Li,
  Yuan, Wang et~al.}]{sun2021sparse}
\bibinfo{author}{Sun, P.}, \bibinfo{author}{Zhang, R.}, \bibinfo{author}{Jiang,
  Y.}, \bibinfo{author}{Kong, T.}, \bibinfo{author}{Xu, C.},
  \bibinfo{author}{Zhan, W.}, \bibinfo{author}{Tomizuka, M.},
  \bibinfo{author}{Li, L.}, \bibinfo{author}{Yuan, Z.}, \bibinfo{author}{Wang,
  C.} et~al. (\bibinfo{year}{2021}).
\newblock \bibinfo{title}{Sparse r-cnn: End-to-end object detection with
  learnable proposals}.
\newblock In {\it \bibinfo{booktitle}{Proceedings of the IEEE/CVF conference on
  computer vision and pattern recognition}\/} (pp.
  \bibinfo{pages}{14454--14463}).
\bibitem[{Tian et~al.(2020{\natexlab{a}})Tian, Lin, Zhou, Duan, Cao, Zhao \&
  Cao}]{tian2020sa}
\bibinfo{author}{Tian, D.}, \bibinfo{author}{Lin, C.}, \bibinfo{author}{Zhou,
  J.}, \bibinfo{author}{Duan, X.}, \bibinfo{author}{Cao, Y.},
  \bibinfo{author}{Zhao, D.}, \& \bibinfo{author}{Cao, D.}
  (\bibinfo{year}{2020}{\natexlab{a}}).
\newblock \bibinfo{title}{Sa-yolov3: An efficient and accurate object detector
  using self-attention mechanism for autonomous driving}.
\newblock {\it \bibinfo{journal}{IEEE Transactions on Intelligent
  Transportation Systems}\/}, .
\bibitem[{Tian et~al.(2020{\natexlab{b}})Tian, Shen, Chen \& He}]{tian2020fcos}
\bibinfo{author}{Tian, Z.}, \bibinfo{author}{Shen, C.}, \bibinfo{author}{Chen,
  H.}, \& \bibinfo{author}{He, T.} (\bibinfo{year}{2020}{\natexlab{b}}).
\newblock \bibinfo{title}{Fcos: A simple and strong anchor-free object
  detector}.
\newblock {\it \bibinfo{journal}{IEEE Transactions on Pattern Analysis and
  Machine Intelligence}\/}, .
\bibitem[{Vahl et~al.(2021)Vahl, Gutsche, Bestmann \& Zhang}]{vahl2021yoeo}
\bibinfo{author}{Vahl, F.}, \bibinfo{author}{Gutsche, J.},
  \bibinfo{author}{Bestmann, M.}, \& \bibinfo{author}{Zhang, J.}
  (\bibinfo{year}{2021}).
\newblock \bibinfo{title}{Yoeo--you only encode once: A cnn for embedded object
  detection and semantic segmentation}.
\newblock In {\it \bibinfo{booktitle}{2021 IEEE International Conference on
  Robotics and Biomimetics (ROBIO)}\/} (pp. \bibinfo{pages}{619--624}).
\newblock \bibinfo{organization}{IEEE}.
\bibitem[{Wang et~al.(2021{\natexlab{a}})Wang, Bochkovskiy \&
  Liao}]{wang2021scaled}
\bibinfo{author}{Wang, C.-Y.}, \bibinfo{author}{Bochkovskiy, A.}, \&
  \bibinfo{author}{Liao, H.-Y.~M.} (\bibinfo{year}{2021}{\natexlab{a}}).
\newblock \bibinfo{title}{Scaled-yolov4: Scaling cross stage partial network}.
\newblock In {\it \bibinfo{booktitle}{Proceedings of the IEEE/cvf conference on
  computer vision and pattern recognition}\/} (pp.
  \bibinfo{pages}{13029--13038}).
\bibitem[{Wang et~al.(2022{\natexlab{a}})Wang, Bochkovskiy \&
  Liao}]{wang2022yolov7}
\bibinfo{author}{Wang, C.-Y.}, \bibinfo{author}{Bochkovskiy, A.}, \&
  \bibinfo{author}{Liao, H.-Y.~M.} (\bibinfo{year}{2022}{\natexlab{a}}).
\newblock \bibinfo{title}{Yolov7: Trainable bag-of-freebies sets new
  state-of-the-art for real-time object detectors}.
\newblock {\it \bibinfo{journal}{arXiv preprint arXiv:2207.02696}\/}, .
\bibitem[{Wang et~al.(2020)Wang, Liao, Wu, Chen, Hsieh \& Yeh}]{wang2020cspnet}
\bibinfo{author}{Wang, C.-Y.}, \bibinfo{author}{Liao, H.-Y.~M.},
  \bibinfo{author}{Wu, Y.-H.}, \bibinfo{author}{Chen, P.-Y.},
  \bibinfo{author}{Hsieh, J.-W.}, \& \bibinfo{author}{Yeh, I.-H.}
  (\bibinfo{year}{2020}).
\newblock \bibinfo{title}{Cspnet: A new backbone that can enhance learning
  capability of cnn}.
\newblock In {\it \bibinfo{booktitle}{Proceedings of the IEEE/CVF conference on
  computer vision and pattern recognition workshops}\/} (pp.
  \bibinfo{pages}{390--391}).
\bibitem[{Wang et~al.(2018)Wang, Chen, Yuan, Liu, Huang, Hou \&
  Cottrell}]{wang2018understanding}
\bibinfo{author}{Wang, P.}, \bibinfo{author}{Chen, P.}, \bibinfo{author}{Yuan,
  Y.}, \bibinfo{author}{Liu, D.}, \bibinfo{author}{Huang, Z.},
  \bibinfo{author}{Hou, X.}, \& \bibinfo{author}{Cottrell, G.}
  (\bibinfo{year}{2018}).
\newblock \bibinfo{title}{Understanding convolution for semantic segmentation}.
\newblock In {\it \bibinfo{booktitle}{2018 IEEE winter conference on
  applications of computer vision (WACV)}\/} (pp. \bibinfo{pages}{1451--1460}).
\newblock \bibinfo{organization}{Ieee}.
\bibitem[{Wang et~al.(2021{\natexlab{b}})Wang, Xie, Li, Fan, Song, Liang, Lu,
  Luo \& Shao}]{wang2021pyramid}
\bibinfo{author}{Wang, W.}, \bibinfo{author}{Xie, E.}, \bibinfo{author}{Li,
  X.}, \bibinfo{author}{Fan, D.-P.}, \bibinfo{author}{Song, K.},
  \bibinfo{author}{Liang, D.}, \bibinfo{author}{Lu, T.}, \bibinfo{author}{Luo,
  P.}, \& \bibinfo{author}{Shao, L.} (\bibinfo{year}{2021}{\natexlab{b}}).
\newblock \bibinfo{title}{Pyramid vision transformer: A versatile backbone for
  dense prediction without convolutions}.
\newblock In {\it \bibinfo{booktitle}{Proceedings of the IEEE/CVF International
  Conference on Computer Vision}\/} (pp. \bibinfo{pages}{568--578}).
\bibitem[{Wang et~al.(2022{\natexlab{b}})Wang, Xie, Li, Fan, Song, Liang, Lu,
  Luo \& Shao}]{wang2022pvt}
\bibinfo{author}{Wang, W.}, \bibinfo{author}{Xie, E.}, \bibinfo{author}{Li,
  X.}, \bibinfo{author}{Fan, D.-P.}, \bibinfo{author}{Song, K.},
  \bibinfo{author}{Liang, D.}, \bibinfo{author}{Lu, T.}, \bibinfo{author}{Luo,
  P.}, \& \bibinfo{author}{Shao, L.} (\bibinfo{year}{2022}{\natexlab{b}}).
\newblock \bibinfo{title}{Pvt v2: Improved baselines with pyramid vision
  transformer}.
\newblock {\it \bibinfo{journal}{Computational Visual Media}\/},  {\it
  \bibinfo{volume}{8}\/}, \bibinfo{pages}{415--424}.
\bibitem[{Yang et~al.(2020)Yang, Zhang, Wang, Xu, Deng \& Yang}]{yang2020part}
\bibinfo{author}{Yang, P.}, \bibinfo{author}{Zhang, G.}, \bibinfo{author}{Wang,
  L.}, \bibinfo{author}{Xu, L.}, \bibinfo{author}{Deng, Q.}, \&
  \bibinfo{author}{Yang, M.-H.} (\bibinfo{year}{2020}).
\newblock \bibinfo{title}{A part-aware multi-scale fully convolutional network
  for pedestrian detection}.
\newblock {\it \bibinfo{journal}{IEEE Transactions on Intelligent
  Transportation Systems}\/},  {\it \bibinfo{volume}{22}\/},
  \bibinfo{pages}{1125--1137}.
\bibitem[{Ye et~al.(2020{\natexlab{a}})Ye, Ren, Zhang, Zhai \&
  Wang}]{ye2020application}
\bibinfo{author}{Ye, T.}, \bibinfo{author}{Ren, C.}, \bibinfo{author}{Zhang,
  X.}, \bibinfo{author}{Zhai, G.}, \& \bibinfo{author}{Wang, R.}
  (\bibinfo{year}{2020}{\natexlab{a}}).
\newblock \bibinfo{title}{Application of lightweight railway transit object
  detector}.
\newblock {\it \bibinfo{journal}{IEEE Transactions on Industrial
  Electronics}\/},  {\it \bibinfo{volume}{68}\/},
  \bibinfo{pages}{10269--10280}.
\bibitem[{Ye et~al.(2020{\natexlab{b}})Ye, Zhang, Zhang \& Liu}]{ye2020railway}
\bibinfo{author}{Ye, T.}, \bibinfo{author}{Zhang, X.}, \bibinfo{author}{Zhang,
  Y.}, \& \bibinfo{author}{Liu, J.} (\bibinfo{year}{2020}{\natexlab{b}}).
\newblock \bibinfo{title}{Railway traffic object detection using differential
  feature fusion convolution neural network}.
\newblock {\it \bibinfo{journal}{IEEE Transactions on Intelligent
  Transportation Systems}\/},  {\it \bibinfo{volume}{22}\/},
  \bibinfo{pages}{1375--1387}.
\bibitem[{Yu et~al.(2020)Yu, Chen, Wang, Xian, Chen, Liu, Madhavan \&
  Darrell}]{yu2020bdd100k}
\bibinfo{author}{Yu, F.}, \bibinfo{author}{Chen, H.}, \bibinfo{author}{Wang,
  X.}, \bibinfo{author}{Xian, W.}, \bibinfo{author}{Chen, Y.},
  \bibinfo{author}{Liu, F.}, \bibinfo{author}{Madhavan, V.}, \&
  \bibinfo{author}{Darrell, T.} (\bibinfo{year}{2020}).
\newblock \bibinfo{title}{Bdd100k: A diverse driving dataset for heterogeneous
  multitask learning}.
\newblock In {\it \bibinfo{booktitle}{Proceedings of the IEEE/CVF conference on
  computer vision and pattern recognition}\/} (pp.
  \bibinfo{pages}{2636--2645}).
\bibitem[{Zeng et~al.(2022)Zeng, Wu, Wang, Li, Liu \& Liu}]{9720996}
\bibinfo{author}{Zeng, N.}, \bibinfo{author}{Wu, P.}, \bibinfo{author}{Wang,
  Z.}, \bibinfo{author}{Li, H.}, \bibinfo{author}{Liu, W.}, \&
  \bibinfo{author}{Liu, X.} (\bibinfo{year}{2022}).
\newblock \bibinfo{title}{A small-sized object detection oriented multi-scale
  feature fusion approach with application to defect detection}.
\newblock {\it \bibinfo{journal}{IEEE Transactions on Instrumentation and
  Measurement}\/},  {\it \bibinfo{volume}{71}\/}, \bibinfo{pages}{1--14}.
\bibitem[{Zhang et~al.(2021{\natexlab{a}})Zhang, Wang, Dayoub \&
  Sunderhauf}]{zhang2021varifocalnet}
\bibinfo{author}{Zhang, H.}, \bibinfo{author}{Wang, Y.},
  \bibinfo{author}{Dayoub, F.}, \& \bibinfo{author}{Sunderhauf, N.}
  (\bibinfo{year}{2021}{\natexlab{a}}).
\newblock \bibinfo{title}{Varifocalnet: An iou-aware dense object detector}.
\newblock In {\it \bibinfo{booktitle}{Proceedings of the IEEE/CVF Conference on
  Computer Vision and Pattern Recognition}\/} (pp.
  \bibinfo{pages}{8514--8523}).
\bibitem[{Zhang et~al.(2021{\natexlab{b}})Zhang, Fu, Xie, Zhu, Tie \&
  Chen}]{zhang2021global}
\bibinfo{author}{Zhang, W.}, \bibinfo{author}{Fu, C.}, \bibinfo{author}{Xie,
  H.}, \bibinfo{author}{Zhu, M.}, \bibinfo{author}{Tie, M.}, \&
  \bibinfo{author}{Chen, J.} (\bibinfo{year}{2021}{\natexlab{b}}).
\newblock \bibinfo{title}{Global context aware rcnn for object detection}.
\newblock {\it \bibinfo{journal}{Neural Computing and Applications}\/},  {\it
  \bibinfo{volume}{33}\/}, \bibinfo{pages}{11627--11639}.
\bibitem[{Zhang et~al.(2021{\natexlab{c}})Zhang, Lu, Cao, Yang, Jiao \&
  Liu}]{zhang2021vit}
\bibinfo{author}{Zhang, Z.}, \bibinfo{author}{Lu, X.}, \bibinfo{author}{Cao,
  G.}, \bibinfo{author}{Yang, Y.}, \bibinfo{author}{Jiao, L.}, \&
  \bibinfo{author}{Liu, F.} (\bibinfo{year}{2021}{\natexlab{c}}).
\newblock \bibinfo{title}{Vit-yolo: Transformer-based yolo for object
  detection}.
\newblock In {\it \bibinfo{booktitle}{Proceedings of the IEEE/CVF International
  Conference on Computer Vision}\/} (pp. \bibinfo{pages}{2799--2808}).

\end{thebibliography}

\end{document}